\documentclass[11pt,a4paper]{article}
\usepackage{naaclhlt2019}
\usepackage{tempora}
\usepackage{latexsym}

\usepackage{graphicx}
\usepackage[russian, english]{babel}

\usepackage[utf8]{inputenc}
\usepackage[T1,T2A]{fontenc}

\usepackage{todonotes}

\usepackage{url}
\usepackage{multirow}

\aclfinalcopy 

\usepackage{array}
\newcolumntype{L}[1]{>{\raggedright\let\newline\\\arraybackslash\hspace{0pt}}m{#1}}
\newcolumntype{C}[1]{>{\centering\let\newline\\\arraybackslash\hspace{0pt}}m{#1}}
\newcolumntype{R}[1]{>{\raggedleft\let\newline\\\arraybackslash\hspace{0pt}}m{#1}}

\title{Char-RNN for Word Stress Detection in East Slavic Languages}

\author{ Maria Ponomareva \\
  ABBYY  \\
  Moscow, Russia \\
  {\tt maria.ponomareva@abbyy.com} \\\And
  Kirill Milintsevich \\
  University of Tartu  \\
  Tartu, Estonia\\
  {\tt kirill.milintsevich@ut.ee} \\\And 
  Ekaterina Artemova \\
  National Research University  \\
  Higher School of Economics \\
  Moscow, Russia \\
  {\tt echernyak@hse.ru} }

  \date{}

\begin{document}
\maketitle
\begin{abstract}

We explore how well a sequence labeling approach, namely, recurrent neural network, is suited for the task of resource-poor and POS tagging free word stress detection in the Russian, Ukranian, Belarusian languages. We present new datasets, annotated with the word stress, for the three languages and compare several RNN models trained on three languages and explore possible applications of the transfer learning for the task. We show that it is possible to  train a  model in a cross-lingual setting  and that using additional languages improves the quality of the results. \end{abstract}


\section{Introduction}
It is impossible to describe Russian (and any other East Slavic) word stress with a set of hand-picked rules. While the stress can be fixed at a word base or ending along the whole paradigm, it can also change its position. The word stress detection task is important for text-to-speech solutions and word-level homonymy resolving. Moreover, stress detecting software is in demand among Russian learners.

One of the approaches to solving this problem is a dictionary-based system. It simply keeps all the wordforms and fails at OOV-words. The rule-based approach offers better results; however collecting the word stress patterns is a highly time consuming task. Also, the method cannot manage words without special morpheme markers. As shown in \cite{ponomareva2017automated}, even simple deep learning methods easily outperform all the approaches described above. 


In this paper we address the following research questions:
\begin{enumerate}
    \item how well does the sequence labeling approach suit the word stress detection task? 
    \item among the investigated RNN-based architectures, what is the best one for the task?
    \item can a word detection system be trained on one or a combination of languages and successfully used for another language? 
\end{enumerate}
To tackle these questions we:
\begin{enumerate}
    \item compare the investigated RNN-based models for the word stress detection task on a standard dataset in Russian and select the best one;
    \item create new data sets in Russian, Ukrainian and Belarusian and conduct a series of mono- and cross-lingual experiments to study the possibility of cross-lingual analysis. 
\end{enumerate}

The paper is structured as follows: we start with the description of the datasets created. Next, we present our major approach to the selection of neural network architecture. Finally, we discuss the results and related work. 

\section{Dataset}
In this project, we approach the word stress detection problem for three East Slavic languages: Russian, Ukrainian and Belarusian, which are said to be mutually intelligible to some extent. Our preliminary experiments along with the results of \cite{ponomareva2017automated} show that using context, i.e., left and right words to the word under consideration, is of great help. Hence, such data sources as dictionaries, including Wiktionary, do not satisfy these requirements, because they provide only single words and do not provide context words. 

To our knowledge, there are no corpora, annotated with word stress for Ukrainian and Belarusian, while there are available transcriptions from the speech subcorpus in Russian\footnote{Word stress in spoken texts database in Russian National Corpus [Baza dannykh aktsentologicheskoy razmetki ustnykh tekstov v sostave Natsional'nogo korpusa russkogo yazyka], \url{http://www.ruscorpora.ru/en/search-spoken.html}} of Russian National Corpus (RNC) \cite{grishina2003spoken}. Due to the lack of necessary corpora, we decided to create them manually. 

The approach to data annotation is quite simple: we adopt texts from Universal Dependencies project and use provided tokenization and POS-tags, conduct simple filtering and use a crowdsourcing platform, Yandex.Toloka\footnote{\url{https://toloka.yandex.ru}}, for the actual annotation. 

\begin{figure}[h!]
    \centering
    \includegraphics[width =0.4\textwidth]{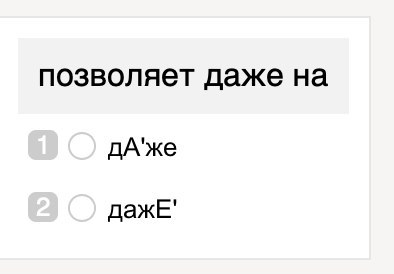}
    \caption{A screenshot of the word stress detection task from Yandex.Toloka crowdsourcing platform}
    \label{fig:toloka}
\end{figure}

To be more precise, we took Russian, Ukrainian and Belarusian treebanks from Universal Dependencies project. We split each text from these treebanks in word trigrams and filtered out unnecessary trigrams, where center words correspond to NUM, PUNCT, and other non-word tokens. The next step is to create annotation tasks for the crowdsourcing platform. We formulate word stress annotation task as a multiple choice task: given a trigram, the annotator has to choose the word stress position in the central word by choosing one of the answer options. Each answer option is the central word, where one of the vowels is capitalized to highlight a possible word stress position. The example of an annotation task is provided in Fig.~\ref{fig:toloka}. Each task was solved by three annotators. As the task is not complicated, we decide to accept only those tasks where all three annotators would agree. Finally, we obtained three sets of trigrams for the Russian, Ukrainian and Belarusian languages of approximately the following sizes 20K, 10K, 3K correspondingly. The sizes of the resulting datasets are almost proportional to the initial corpora from the Universal Dependencies treebanks.

Due to the high quality of the Universal Dependencies treebanks and the languages being not confused, there are little intersections between the datasets, i.e., only around 50 words are shared between Ukranian and Belarusian datasets and between Russian and Ukranian and Belarusian datasets. The intersection between the Ukrainian and Russian datasets amounts around 200 words. 

The structure of the dataset is straightforward: each entry consists of a word trigram and a number, which indicates the position of the word stress in the central word\footnote{Datasets are avaialable at: \url{https://github.com/MashaPo/russtress}}.

\section{Preprocessing}

We followed a basic preprocessing strategy for all the datasets. First, we tokenize all the texts into words. Next, to take the previous and next word into account we define left and right contexts of the word as the last three characters of the previous word and last three characters of the next word. The word stresses (if any) are removed from context characters.
If the previous / next word has less than three letters, we concatenate it with the current word (for example, \textit{``te\_oblak{\'a}''} [that-Pl.Nom cloud-Pl.Nom]). 
This definition of context is used since East Slavic endings are typically two-four letters long and derivational morphemes are usually located on the right periphery of the word.

Finally, each character is annotated with one of the two labels $\mathcal{L} = \{0, 1\}$: it is annotated with 0, if there is no stress, and with 1, if there should be a stress. An example of an input character string can be found in Table~\ref{table:local_char_ex}.
\begin{table}[ht]
\begin{tabular}{|p{0.5cm}||p{0.07cm}p{0.07cm}p{0.07cm}p{0.07cm}p{0.07cm}p{0.07cm}p{0.07cm}p{0.07cm}p{0.08cm}p{0.07cm}p{0.07cm}p{0.07cm}p{0.07cm}p{0.07cm}|}
\hline
in & л & а & я &   & в & о & р & о & н & а  & & т & и & т  \\ \hline
out & 0 & 0 & 0 & 0 & 0 & 0 & 0 & 1 & 0 & 0 & 0 & 0 & 0 & 0  \\
\hline
\end{tabular}
\caption{Character model input and output: each character is annotated with either 0, or 1. A trigram ``белая ворона летит'' (``white crow flies'') is annotated. The central word remains unchanged, while its left and right contexts are reduced to the last three characters}
\label{table:local_char_ex}
\end{table}

\section{Model selection}

We treat word stress detection as a sequence labeling task. Each character (or syllable) is labeled with one of the labels $\mathcal{L} = \{0, 1\}$, indicating no stress on the character  (0) or a stress (1). Given a string $s = {s_1, \ldots, s_n}$ of characters, the task is to find the labels $Y^* = {y_1^*, \ldots, y_n^*}$, such that 
\[ Y^* = \arg \max_{Y \in \mathcal{L} ^n} p(Y | s).\]

The most probable label is assigned to each character.

We compare two RNN-based models for the task of word stress detection (see Fig.~\ref{fig:locm} and Fig.\ref{fig:globm}). Both models have a common input and hidden layers but differ in output layers.

The input of both models are embeddings of the characters. In both cases, we use bidirectional LSTM of 32 units as the hidden layer. Further, we describe the difference between the output layers of the two models.

\subsection{Local model}

\begin{figure}[h!]
\centering
\includegraphics[width=4.5cm]{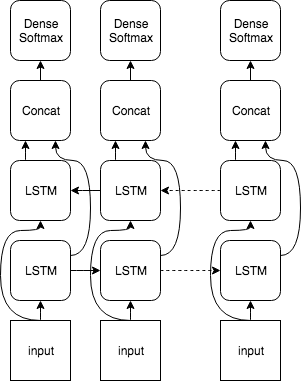}
\caption{Local model for word stress detection}
\label{fig:locm}
\end{figure}

The decision strategy of the local model (see Fig.~\ref{fig:locm}) follows common language modeling and NER architectures \cite{ma2016end}: all outputs are independent of each other. We decide, whether there should be a stress on each given symbol (or syllable) or not. To do this for each character we put an own dense layer with two units and a ${softmax}$ activation function, applied to the corresponding hidden state of the recurrent layer, to label each input character (or syllable) with $\mathcal{L} = \{0, 1\}$. 

\subsection{Global model} 

\begin{figure}[h!]
\centering
\includegraphics[width=4.5cm]{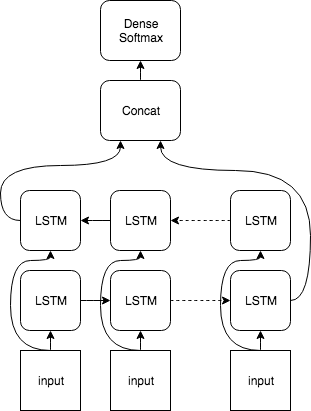}
\caption{Global model for word stress detection}
\label{fig:globm}
\end{figure}

The decision strategy of the global model (see  Fig.~\ref{fig:globm}) follows common encoder-decoder architectures \cite{sutskever2014sequence}. We use the hidden layer to encode the input sequence into a vector representation.
Then, we use a dense layer of $n$ units as a decoder to decode the representation of the input and to generate the desired sequence of $\{0, 1\}$.
In comparison to the local model, in this case, we try to find the position of the stress instead of making a series of local decisions if there should be a stress on each character or not. 

To test the approach and to compare these models, we train two models on the subcorpus of Russian National Corpus for Word stress in spoken texts, which appears to be a standard dataset for the task of word stress detection. This dataset was preprocessed according to our standard procedure, and the resulting dataset contains approximately around 1M trigrams. The results of cross-validation experiments, presented in Table~\ref{table:res_russian}, show that the global model outperforms significantly the local model. Hence, the global architecture is used further on in the next experiments.

\begin{table}[h!]
\centering
\begin{tabular}{|p{2cm}|p{1cm}|p{1cm}|}
\hline

\# vowels & local & global            \\ \hline
& \multicolumn{2}{c|}{all words}           \\ \hline
2 & $961$ &  $983$   \\ \hline
3 & $940$ &  $977$    \\ \hline
4 & $947$ &  $976$      \\ \hline
5 &  $960$ &  $977$     \\ \hline
6 &  $958$ &  $973$      \\ \hline
7 &  $924$ & $955$        \\ \hline
8 &  $866$ & $923$     \\ \hline
9 &  $809$ & $979$      \\ \hline
avg &  $\mathbf{952}$ &  $\mathbf{979}$     \\ \hline
& \multicolumn{2}{c|}{homographs}        \\ \hline
2 &  $839$ &  $810$    \\ \hline
3 &  $774$ &  $844$      \\ \hline
4 &  $787$ &  $847$       \\ \hline
avg &  $821$ &  $819$    \\ \hline
\end{tabular}
\caption{Accuracy scores $\times$ 1000 for two models}
\label{table:res_russian}
\end{table}

We pay special attention to homographs: as one can see, in general, the quality of word stress detection is significantly lower on homographs than on regular words. However, in the majority of cases, we are still able to detect the word stress position for a homograph, most likely due to the understanding of the word context. 

\section{Experiments and results}

In these series of experiments, we tried to check the following assumptions for various experiment settings:
\begin{enumerate}
\item monolingual setting: the presented above approach applies not only to the Russian language word stress detection but also to the other East Slavic languages
\item cross-lingual setting (1): it is possible to train a model on one language (e.g., Ukrainian) and test it on another language (e.g., Belarusian) and achieve results comparable to monolingual setting
\item cross-lingual setting (2):  training on several languages (e.g., Russian and Ukrainian) will improve the results of testing on a single language (e.g., Russian) in comparison to the monolingual setting. 
\end{enumerate}

To conduct the experiments in these mono- and cross-lingual settings, we split the annotated datasets for  Russian, Ukrainian and Belarusian randomly in the 7:3 train-test ratio and conducted 20 runs of training and testing with different random seeds. Afterward, the accuracy scores of all runs were averaged. The Table~\ref{table:res_all} presents the results of these experiments.

\begin{table}[h!]
\begin{tabular}{|L{2.5cm}|| C{1cm}| C{1cm}| C{1cm}|}
\hline

 & \multicolumn{3}{c|}{test dataset} \\ \hline
train dataset & Belarusian & Russian  & Ukrai-nian \\ \hline
Belarusian &  647 & 326 & 373 \\ \hline
Russian &  495 & 738 & 516 \\ \hline
Ukrainian & 556 & 553 & 683 \\ \hline
Ukrainian, Belarusian & 769 & 597 & 701 \\ \hline
Russian, Belarusian & 740 & 740 & 563 \\ \hline
Russian, Ukrainian & 627 & 756 & {\bf 700} \\ \hline
Russian, Ukrainian, Belarusian & {\bf 772} & {\bf 760} & 698 \\ \hline
\end{tabular}
\caption{Accuracy scores $\times$ 1000 for different train and test dataset combinations }
\label{table:res_all}
\end{table}

The Table~\ref{table:res_all} shows, that:
\begin{enumerate}
    \item in monolingual setting, we can get high-quality results. The scores are significantly lower than the scores of the same model on the standard dataset, due to the smaller sizes of the training datasets. Nevertheless, one can see, that our approach to word stress detection applies not only to the Russian language data, but also to the data in the Belarusian and Ukrainian languages;
    \item cross-lingual setting (1): the Belarusian training dataset, being the smallest one among the three datasets, is not a good source for training word stress detection models in other languages, while the Ukrainian dataset stands out as a good source for training word stress detection systems both for the Russian and Belarusian languages; 
    \item cross-lingual setting (2): adding one or two datasets to the other languages improves the quality. For example, around 10\% of accuracy is gained by adding the Russian training dataset to the Belarusian training dataset, while testing on Belarusian. 
\end{enumerate}

One possible reason for the difference of Belarusian from the other two languages can be the following. After the orthography reform in 1933, the cases of vowel reduction in the unstressed position (common phonetic feature for East Slavic languages) have been represented orthographically in the Belarusian language. However, the size of the Belarusian dataset (it is much smaller than the other two) may affect the quality as well.

\section{Related Work}
\label{sec:sec5}

\subsection{Char-RNN models}
\label{ssec:subhead1}

Several research groups have shown that character-level models are an efficient way to deal with unseen words in various NLP tasks, such as text classification \cite{joulin2017bag}, named entity recognition \cite{ma2016end}, POS-tagging \cite{santos2014learning,cotterell2017cross}, dependency parsing \cite{alberti2017syntaxnet} or machine translation \cite{chungcharacter}. The character-level model is a model which either treats the text as a sequence of characters without any tokenization or incorporates character-level information into word-level information. Character-level models can capture morphological patterns, such as prefixes and suffixes so that the model can define the POS-tag or NE class of an unknown word. 

\subsection{Word stress detection in East Slavic languages}

Only a few authors touch upon the problem of automated word stress detection in Russian. Among them, one research project, in particular, is worth mentioning \cite{hall-sproat2013EMNLP}. The authors restricted the task of stress detection to find the correct order within an array of stress assumptions where valid stress patterns were closer to the top of the list than the invalid ones. Then, the first stress assumption in the rearranged list was considered to be correct. The authors used the Maximum Entropy Ranking method to address this problem \cite{collins2005russian} and took character bi- and trigram, suffixes and prefixes of ranked words as features as well as suffixes and prefixes represented in an ``abstract'' form where most of the vowels and consonants were replaced with their phonetic class labels. The study features the results obtained using the corpus of Russian wordforms generated based on Zaliznyak’s Dictionary (approx. 2m wordforms). Testing the model on a randomly split train and test samples showed the accuracy of 0.987. According to the authors, they observed such a high accuracy because splitting the sample randomly during testing helped the algorithm benefit from the lexical information, i.e., different wordforms of the same lexical item often share the same stress position. The authors then tried to solve a more complicated problem and tested their solution on a small number of wordforms for which the paradigms were not included in the training sample. As a result, the accuracy of 0.839 was achieved. The evaluation technique that the authors propose is quite far from a real-life application which is the main disadvantage of their study. Usually, the solutions in the field of automated stress detection are applied to real texts where the frequency distribution of wordforms differs drastically from the one in a bag of words obtained from ``unfolding'' of all the items in a dictionary.

Also, another study \cite{reynolds-tyers2015NODALIDA} describes the rule-based method of automated stress detection without the help of machine learning. The authors proposed a system of finite-state automata imitating the rules of Russian stress accentuation and formal grammar that partially solved stress ambiguity by applying syntactical restrictions. Thus, using all the above-mentioned solutions together with wordform frequency information, the authors achieved the accuracy of 0.962 on a relatively small hand-tagged Russian corpus (7689 tokens) that was not found to be generally available. We can treat the proposed method as a baseline for the automated word stress detection problem in Russian.

The global model, which is shown to be the best RNN-based architecture for this setting of the task, was first presented in \cite{ponomareva2017automated}, where a simple bidirectional RNN with LSTM nodes was used to achieve the accuracy of 90\% or higher. The authors experiment with two training datasets and show that using the data from an annotated corpus is much more efficient than using a dictionary since it allows to consider word frequencies and the morphological context of the word. We extend the approach of \cite{ponomareva2017automated} by training on new datasets from additional languages and conducting cross-lingual experiments.

\subsection{Cross-lingual analysis}
Cross-lingual analysis has received some attention in the NLP community, especially when applied in neural systems. Among a few research directions of cross-lingual analysis are multilingual word embeddings \cite{ammar2016massively,hermann2013multilingual} and dialect identification systems \cite{malmasi2016discriminating,al2015aida2}. Traditional NLP tasks such as POS-tagging \cite{cotterell2017cross}, morphological reinflection \cite{kann2017one} and dependency parsing \cite{guo2015cross} benefit from cross-lingual training too. Although the above-mentioned tasks are quite diverse, the undergirding philosophical motivation is similar: to approach a task on a low-resource language by using additional training data in a high-resource language or training a model on a high-resource language and fine-tune this model on a low-resource language with a probably lower learning rate.

\section{Conclusion}

In this project, we present a neural approach for word stress detection. We test the approach in several settings: first, we compare several neural architectures on a standard dataset for the Russian language and use the results of this experiment to select the architecture that provides the highest accuracy score. 
Next, we annotated the Universal Dependencies corpora for the Russian, Ukrainian and Belarusian languages with word stress using Yandex. Toloka crowdsourcing platform. The experiments conducted on these datasets consist of two parts: a) in the monolingual setting we train and test the model for word stress detection on the data sets separately; b) in the cross-lingual setting: we train the model on various combinations of the datasets and test on all three data sets.
These experiments show that:
\begin{enumerate}
    \item the proposed method for word stress detection is applicable or the Russian, Ukrainian and Belarusian languages;
    \item using an additional language for training most likely improves the quality of the results.
\end{enumerate}
Future work should focus on both annotating new datasets for other languages that possess word stress phenomena and further development of cross-lingual neural models based on other sequence processing architectures, such as transformers.

\bibliography{refs}
\bibliographystyle{acl_natbib}

\end{document}